\documentclass[lettersize,journal]{IEEEtran}
\usepackage{amsmath,amsfonts}
\usepackage{algorithmic}
\usepackage{algorithm}
\usepackage{array}
\usepackage[caption=false,font=normalsize,labelfont=sf,textfont=sf]{subfig}
\usepackage{textcomp}
\usepackage{stfloats}
\usepackage{url}
\usepackage{verbatim}
\usepackage{graphicx}
\usepackage{cite}
\usepackage{bbding}
\usepackage{pifont}
\usepackage{array}
\usepackage{svg}
\usepackage{makecell}
\usepackage{amsmath}
\usepackage{multirow}
\usepackage{enumitem}
\usepackage{tabularx}
\usepackage{soul}

\usepackage[colorlinks,
linkcolor=blue,
anchorcolor=blue,
citecolor=blue]{hyperref}
\usepackage{orcidlink}

\begin{document}

\title{Bring Remote Sensing Object Detect Into Nature Language Model: Using SFT Method}

\author{Fei Wang\orcidlink{0009-0005-3061-0496}, Chengcheng Chen\orcidlink{0000-0002-6014-5135},~\IEEEmembership{Graduate Student Member, IEEE}, \\
	Hongyu Chen\orcidlink{0009-0004-2934-3135}, Yugang Chang\orcidlink{0009-0009-9247-8098}, and Weiming Zeng\orcidlink{0000-0002-9035-8078},~\IEEEmembership{Senior Member, IEEE}
	\thanks{This paragraph of the first footnote will contain the date on which you submitted your paper for review. It will also contain support information, including sponsor and financial support acknowledgment. This work was supported by the National Natural Science Foundation of China (grant nos. 31870979), in part by the 2023 Graduate Top Innovative Talents Training Program at Shanghai Maritime University under Grant 2023YBR013.  (Corresponding author: Weiming Zeng)}
	\thanks{The authors are with the Digital Imaging and Intelligent Computing Laboratory, Shanghai Maritime University, Shanghai 201306, China. (E-mail: shine\_wxf@163.com;  shmtu\_ccc@163.com; hongychen676@gmail.com; cygang\_post@163.com; zengwm86@163.com)}}


\maketitle

\begin{abstract}
Recently, large language models (LLMs) and vision-language models (VLMs) have achieved significant success, demonstrating remarkable capabilities in understanding various images and videos, particularly in classification and detection tasks. However, due to the substantial differences between remote sensing images and conventional optical images, these models face considerable challenges in comprehension, especially in detection tasks. Directly prompting VLMs with detection instructions often leads to unsatisfactory results. To address this issue, this letter explores the application of VLMs for object detection in remote sensing images. Specifically, we constructed supervised fine-tuning (SFT) datasets using publicly available remote sensing object detection datasets, including SSDD, HRSID, and NWPU-VHR-10. In these new datasets, we converted annotation information into JSON-compliant natural language descriptions, facilitating more effective understanding and training for the VLM. We then evaluate the detection performance of various fine-tuning strategies for VLMs and derive optimized model weights for object detection in remote sensing images. Finally, we evaluate the model’s prior knowledge capabilities using natural language queries. Experimental results demonstrate that, without modifying the model architecture, remote sensing object detection can be effectively achieved using natural language alone. Additionally, the model exhibits the ability to perform certain vision question answering (VQA) tasks. Our datasets and related code will be released soon.

\end{abstract}

\begin{IEEEkeywords}
Remote Sensing, Object Detect, Vision Language Model(VLM), NLP.
\end{IEEEkeywords}

\section{Introduction}
\IEEEPARstart{C}{urrently}, most object detection models in the field of remote sensing imagery are based on CNN and Transformer\cite{Vaswani2017AttentionIA} architectures and are specifically designed for detection tasks. Examples include Faster R-CNN\cite{7485869}, YOLO\cite{tian2025yolov12}, and DETR\cite{Carion2020EndtoEndOD}, as well as algorithms tailored for remote sensing or SAR images, such as $\rm CS^{n}$Net\cite{10242028} and RSNet\cite{Chen2024RSNetAL}. Although these models have achieved notable success, they are typically limited to single-task detection and lack generalization capabilities. They cannot comprehend user instructions in natural language or perform more complex tasks, such as scene description.

Large language models (LLMs) based on the Transformer architecture have demonstrated remarkable capabilities in understanding human natural language, excelling in tasks such as contextual dialogue, question answering, and logical reasoning. Notable examples include the Qwen\cite{Yang2024Qwen25TR}, GPT\cite{Achiam2023GPT4TR}, and DeepSeek\cite{DeepSeekAI2025DeepSeekR1IR}. With the continuous advancement of these models, researchers have integrated image encoders—such as Vision Transformers—with LLMs, leading to the development of vision-language models (VLMs) that can comprehend visual information. Recently, Qwen2.5-VL\cite{Bai2025Qwen25VLTR} has demonstrated impressive performance across various vision tasks, showcasing its potential for more challenging applications such as object detection. However, due to the differences between remote sensing images and natural images—particularly in the case of SAR images, which contain only black and white tones—the model, despite its extensive visual knowledge, struggles to effectively interpret remote sensing imagery.

Many researchers have started developing language models tailored for remote sensing imagery, bringing remote sensing tasks into the era of natural language models. For instance, Zhang et al. \cite{10547418} introduced EarthGPT, a model based on Llama2\cite{Touvron2023Llama2O}, which utilizes parallel CNN and Vision Transformer (ViT) \cite{DBLP:journals/corr/abs-2010-11929} to extract image features. By training on the MMRS-1M dataset, EarthGPT is capable of performing various remote sensing tasks, such as object detection and visual question answering. However, the parallel use of ViT and CNN for visual feature extraction introduces some redundancy in the model. Wang et al. \cite{10555327} proposed RSAdapter, which focuses on runtime and parameter efficiency, achieving promising results in vision question answering (VQA) tasks within the remote sensing domain. However, this approach is limited to answering a fixed set of questions, as the total number of queries is predetermined during training, restricting its applicability to more complex vision tasks. Additionally, Zhang et al.\cite{10738390} proposed Popeye, a VLM-based approach for ship detection in remote sensing imagery. While this method demonstrates effectiveness for ship detection, its performance on non-ship categories remains unexplored.

Although the aforementioned models have demonstrated strong performance in specific tasks, our goal is to develop an end-to-end VLM for remote sensing object detection without incorporating any modules specifically tailored for remote sensing imagery. This approach enables both conversational capabilities and object detection within a unified framework. In this letter, we fine-tune Qwen2.5-VL-7B to obtain RS-OD-Qwen, a model capable of performing remote sensing object detection using natural language instructions. Additionally, it supports VQA, such as image scene description, further integrating NLP capabilities into remote sensing applications. The letter’s primary contributions are as follows:
\begin{enumerate}[label=\arabic*)]
\item We construct instruction-tuning (SFT) datasets for VLM-based object detection using three remote sensing image datasets: SSDD\cite{rs13183690}, HRSID\cite{9127939}, and NWPU-VHR-10\cite{CHENG2014119}.

\item We conduct fine-tuning experiments on these three datasets and evaluate the model using relevant performance metrics. Our results demonstrate that remote sensing object detection can be achieved without modifying the general VLM architecture. Additionally, the model retains certain VQA capabilities, such as image description.
\end{enumerate}

\begin{figure*}[htbp]
	\centering
	\includegraphics[scale=0.35]{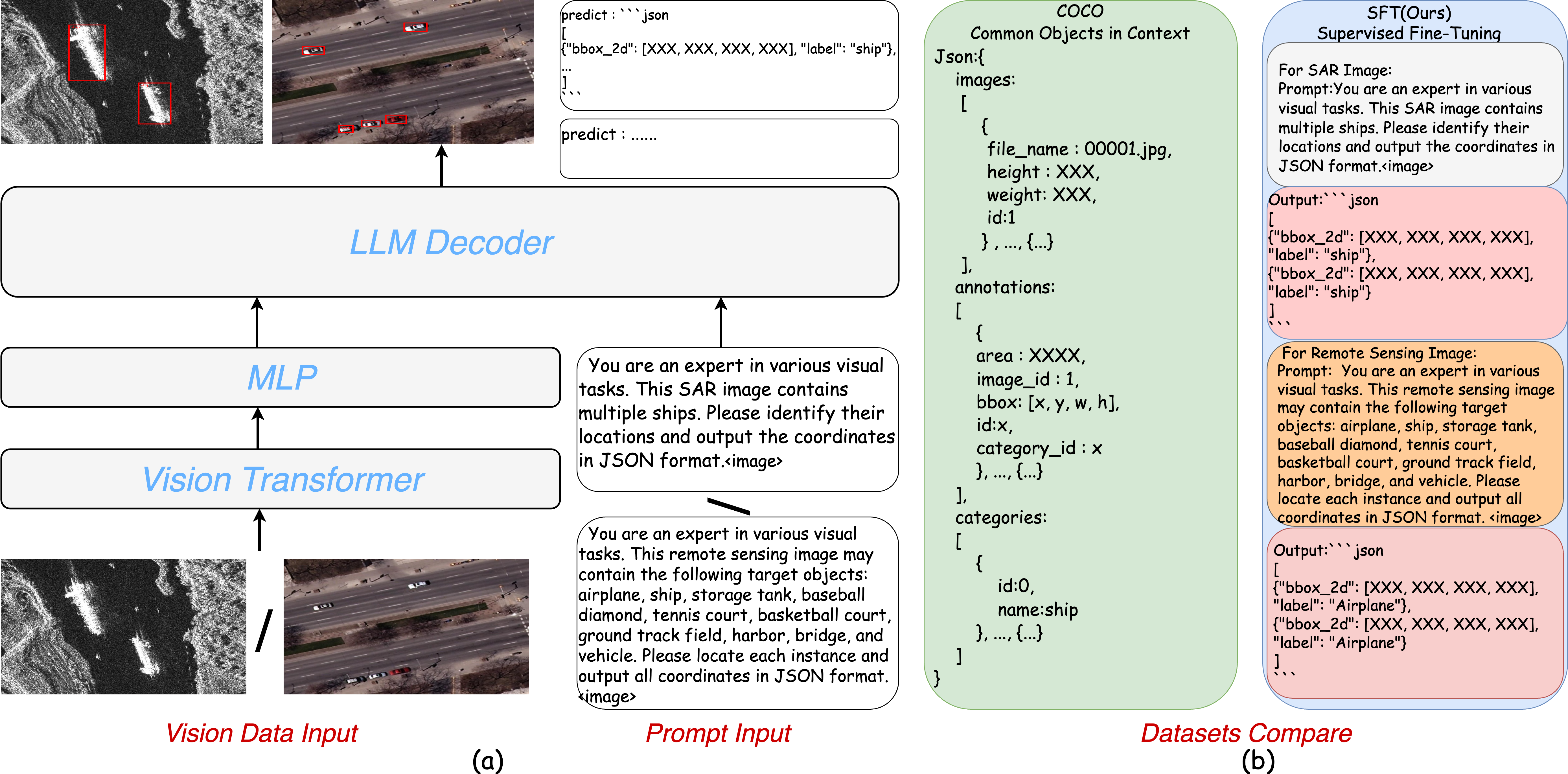}
	\captionsetup{font=footnotesize}
	\caption{Model and dataset format diagram. (a) Model architecture of Qwen2.5-vl. (b) Datasets Compare.}
	\label{img1}
\end{figure*}

\section{Proposed Method}
\subsection{Dataset Construction}
To enable the model to perform object detection via natural language, we construct the dataset following the approach illustrated in Fig. \ref{img1}. Specifically, we introduce prompting instructions to guide the VLM in performing the detection task. The placeholder \textless image\textgreater  represents the input image, and in the label section, we adopt a standardized format to ensure consistency in the model's output and facilitate subsequent coordinate extraction and analysis. We convert the annotations from SSDD, HRSID, and NWPU-VHR-10 into a structured JSON-based natural language format, allowing the model to learn to output ship bounding boxes in a normalized manner. Specifically, as shown in Fig. \ref{img1}(b) , for SAR image data, the input instruction is "You are an expert in various visual tasks. This SAR image contains multiple ships. Please identify their locations and output the coordinates in JSON format. \textless image\textgreater". The output, which serves as the model’s response, adheres to a standardized JSON format to ensure consistency. Each detected object consists of two key components: (1) bbox\_2d: A list in the format [x1, y1, x2, y2], representing the top-left and bottom-right coordinates of the bounding box; (2) label: The category of the detected object. And for optical remote sensing images from the NWPU-VHR-10 dataset, the instruction and output are similar. For SSDD and HRSID, we adhere to the official dataset splits for training and testing to ensure fair performance comparison. For NWPU-VHR-10, we randomly divide the 650 annotated images into an 8:2 training-to-test split. We resize these datasets to further reduce training uncertainty, specifically, following the Qwen2.5-VL paper, we resize all images to dimensions that are multiples of 28, setting them to 644×644 pixels to ensure consistency.
\subsection{Model Fine Tuning}
To maximize the model's potential, we investigate the performance of LoRA fine-tuning under various parameter strategies. This approach enhances the base model by incorporating a parallel low-rank matrix into specific layers, enabling the model to acquire new knowledge with fewer parameters. We explore the model's performance under different rank values to determine the optimal configuration. Additionally, research suggests that adding random noise to the embedding vectors during LLM training can improve fine-tuning results and enhance the model’s response performance \cite{Jain2023NEFTuneNE}. Therefore, we introduce random noise during training. Specifically, let $X_{emb}$ denote the input embedding vector of the model, with a shape of (B, L, d), where B represents the batch size, L the sequence length, and d the embedding dimension. The noise is added according to the following formula:
\begin{equation}
	\label{deqn_ex1}
	X_{emb}^{'}  = X_{emb} + (\frac{\alpha}{\sqrt{Ld}})\epsilon
\end{equation}
Here, $\alpha$ is a hyperparameter representing the base noise scale, and $\epsilon$ is a distribution:
\begin{equation}
	\label{deqn_ex2}
	\epsilon\sim Uniform[-1, 1]
\end{equation}
\subsection{Verify Model Performance Under Different Rank Fine-tuning Strategies}
\begin{figure}[h!]
	\centering
	\includegraphics[scale=0.5]{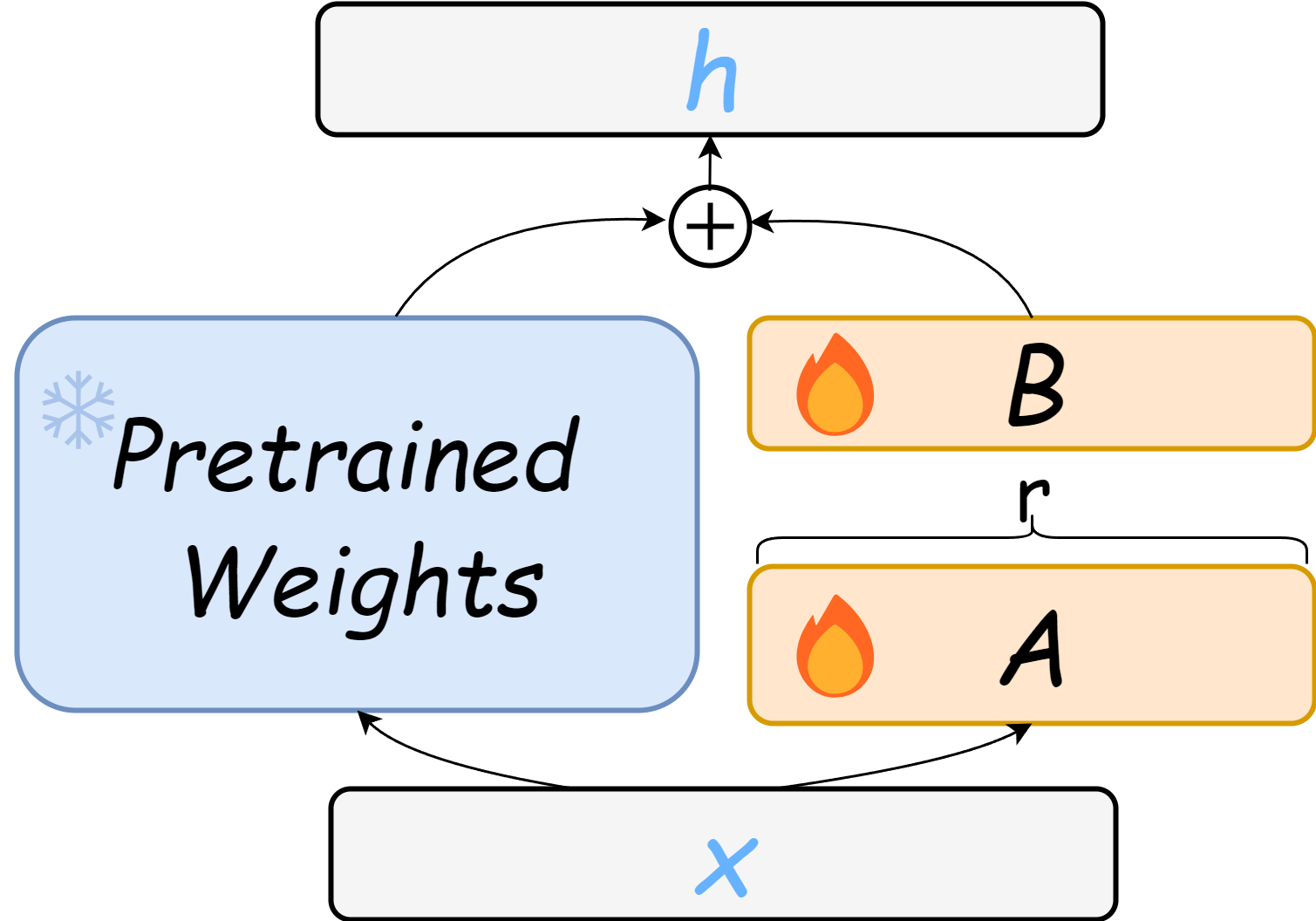}%
	\captionsetup{font=footnotesize}
	\caption{Lora Fine Tuning.}
	\label{img2}
\end{figure}

We employ LoRA fine-tuning \cite{hu2021loralowrankadaptationlarge}, as shown in Fig. \ref{img2}. This method involves inserting a low-rank matrix in parallel into specific pretrained layers of the model, enabling fine-tuning with fewer parameters. For the pre-trained model's weight matrix, $W_0 \in \mathbb{R}^{d \times  k}$, we constrain its updates by introducing a low-rank decomposition $\Delta W=BA$, where $B \in \mathbb{R}^{d \times  r}$ and $A \in \mathbb{R}^{r \times  k}$, with the rank $r \ll min(d, k)$. During training, $W_0$ is frozen and does not receive any gradient updates; only $A$ and $B$ are updated. As shown in Formula \ref{deqn_ex3}, let $h$ denote the output and $x$ be the input. The entire computation process can be expressed as:
\begin{equation}
	\label{deqn_ex3}
	h = W_{0}x + \bigtriangleup Wx = W_{0}x + BAx
\end{equation}

This approach reduces memory consumption and accelerates the training process. During the inference phase, the weights of the pre-trained layers are merged with the LoRA weights to obtain the final model weights. As a result, this approach incurs no additional computational costs during inference.

In summary, through the above process, by optimizing the low-rank matrix $\Delta W$, we introduce LLM’s understanding of remote sensing images, enabling object detection capabilities and producing outputs in a structured format.

\section{Experiments}

\subsection{Implementation Details}
We utilized SSDD, HRSID, and NWPU-VHR-10 as benchmark datasets to construct the corresponding instruction-tuning dataset for natural language fine-tuning. Each instruction data in the datasets includes an image, an input prompt, and a label. We set the number of epochs to 3, the learning rate to 1e-4, and employed a cosine learning rate scheduler with a warm-up ratio of 0.1.

\begin{table}[h!]
	\begin{center}
		\caption{Ablation Experiments on SSDD-Instruct.}
		\label{tab1}
		\begin{tabular}{c | c c c c}
			\Xhline{2\arrayrulewidth}
			\rule{0pt}{0.9em}Rank & P & R & F1-Score  \\
			\hline
			\rule{0pt}{1em}8 & 0.7859 & 0.7125 & 0.7474\\
			\rule{0pt}{1em}16 & 0.8739 & 0.7363 & 0.7992 \\
			\rule{0pt}{1em}32 & 0.8929 & 0.7326 & 0.8048 \\
			\Xhline{2\arrayrulewidth}
		\end{tabular}
	\end{center}
\end{table}

\subsection{Ablation Study}
We validated the performance of LoRA fine-tuning under different ranks on the SSDD-instruct dataset. Additionally, we tested the model's performance with the addition of NEFT noise, as shown in Table \ref{tab1}. For evaluation, we used Precision (P), Recall (R), and F1-Score. It is important to note that these metrics were computed with an IoU threshold of 0.5. Furthermore, if the model’s output JSON could not be correctly parsed, all targets in that image were considered prediction failures and included in the metric calculation, even if part of the model’s output might have been correct. Since the language model directly outputs object bounding boxes through natural language without any confidence scores, the Average Precision (AP) metric is not applicable in this scenario. Finally, based on the experimental results, we selected the training approach with a rank of 32. This configuration allows for updating more parameters, thereby achieving a higher F1-Score.

\subsection{Performance on Different Datasets}
\begin{table}[h!]
	\begin{center}
		\caption{Experimental results based on SAR image training set.}
		\label{tab2}
		\begin{tabular}{c | c c c}
			\Xhline{2\arrayrulewidth}
			\rule{0pt}{0.9em} Dataset &  P& R & F1-Score\\
			\hline
			\rule{0pt}{1em}SSDD & 0.8929 & 0.7326 & 0.8048 \\
			\rule{0pt}{1em}HRSID* (Zero-shot) & 0.7373 & 0.4374 & 0.5490 \\
			\rule{0pt}{1em}HRSID & 0.5930 & 0.5144 & 0.5509 \\
			\Xhline{2\arrayrulewidth}
		\end{tabular}
	\end{center}
	\footnotesize
	* Denotes zero shot based on SSDD training set.
\end{table}

\begin{table}[h!]
	\begin{center}
		\caption{EXPERIMENTS ON NWPU-VHR-10.}
		\label{tab3}
		\begin{tabular}{c | c c c}
			\Xhline{2\arrayrulewidth}
			\rule{0pt}{0.9em} Category&  P& R & F1-Score\\
			\hline
			\rule{0pt}{1em}airplane & 0.8160 & 0.6497 & 0.7234 \\
			\rule{0pt}{1em}baseball diamond & 0.9167 & 0.8871 & 0.9016 \\
			\rule{0pt}{1em}storage tank & 0.2656 & 0.2787 & 0.2720 \\
			\rule{0pt}{1em}ship & 0.4898 & 0.8276 & 0.6154 \\
			\rule{0pt}{1em}ground track field & 0.9268 & 1.0000 & 0.9620 \\
			\rule{0pt}{1em}tennis court & 0.4161 & 0.5135 & 0.4597 \\
			\rule{0pt}{1em}harbor & 0.4237 & 0.3906 & 0.4065 \\
			\rule{0pt}{1em}bridge & 0.3333 & 0.3103 & 0.3214 \\
			\rule{0pt}{1em}vehicle & 0.8161 & 0.7634 & 0.7889 \\
			\rule{0pt}{1em}all & 0.6035 & 0.6062 & 0.6049 \\
			\Xhline{2\arrayrulewidth}
		\end{tabular}
	\end{center}
\end{table}
We conducted additional experiments on HRSID-instruction and NWPU-VHR-10-instruction. In Table \ref{tab2}, we trained the model on SSDD-instruction and evaluated its performance on both the SSDD-instruction and HRSID-instruction datasets. In Table \ref{tab3}, we trained and tested the model on the NWPU-VHR-10-instruction dataset. The experimental results demonstrate that for SAR image ship detection, the model performs reasonably well in a zero-shot experiment on the HRSID-instruction dataset after training on the SSDD-instruction dataset, although the recall rate remains relatively low. This indicates that the model can transfer some learned knowledge from SSDD-instruction to the HRSID-instruction dataset, but it struggles to detect all targets effectively. After fine-tuning the model on HRSID-instruction, both the recall rate and F1-score improve, indicating that training on the target domain (HRSID-instruction) enhances performance. However, this improvement in recall is accompanied by an increase in the total number of detected targets, which may be influenced by noise in the data. As a result, the increased number of detected targets leads to more false positives, resulting in a decrease in precision.

For NWPU-VHR-10, the fine-tuned model achieved strong performance on certain categories, such as airplane, baseball diamond, ground track field, and vehicle. However, due to the limited sample size of this dataset, the model exhibited poor performance on some categories, resulting in lower overall metrics.

\subsection{VQA Experiment}
\begin{figure*}[htbp]
	\centering
	\includegraphics[scale=0.41]{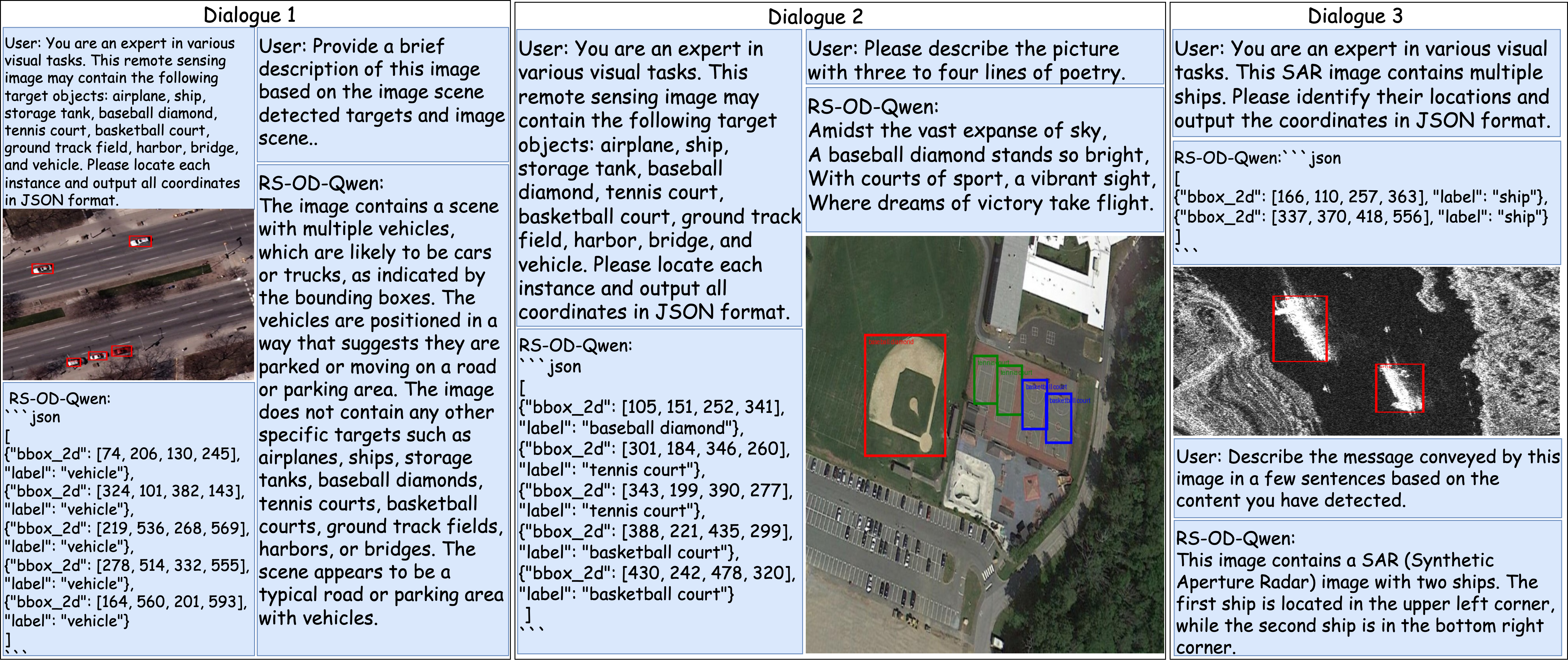} 
	\captionsetup{font=footnotesize}
	\caption{VQA Dialogue.}
	\label{img3}
\end{figure*}
We conducted a VQA dialogue experiment, as shown in Fig. \ref{img3}. In the three dialogues, we first instructed the model to detect objects in the images, followed by a second round of conversation. In Dialogues 1 and 3, we instructed the model to describe the scene in the image, with the first image being an optical remote sensing image and the third a SAR image. The model performed effectively in both descriptions. In Dialogue 2, we instructed the model to compose a short poem describing the image. TThe model's performance demonstrates that fine-tuning not only enabled object detection capabilities but also retained a degree of general knowledge, allowing the model to perform effectively in tasks beyond detection.

\section{Conclusion and Discussion}
Conclusion: In this letter, we construct three  instruction datasets for VLM-based object detection fine-tuning using publicly available remote sensing datasets. The dataset adheres to a standardized JSON format to facilitate data analysis and metric extraction. Additionally, experimental results demonstrate that fine-tuning the model without modifying its architecture enables the acquisition of natural language remote sensing object detection capabilities. The model also supports multi-turn dialogues and certain VQA tasks.

Discussion: Although fine-tuning enables VLMs to acquire object detection capabilities, certain limitations remain. For instance, this study primarily focuses on object detection and image description, while other tasks, such as visual grounding and object counting, are not addressed. In the future, we plan to develop a new dataset incorporating visual grounding and object counting to support region-specific object detection and counting functionalities. These enhancements will further improve the model’s understanding of remote sensing images, thereby increasing its practical applicability. Additionally, we plan to further evaluate the efficiency of VLMs in real-world scenarios and explore methods to optimize inference speed for improved performance.

 
%

\bibliographystyle{IEEEtran}

\bibliography{ref}

\begin{thebibliography}{10}
\providecommand{\url}[1]{#1}
\csname url@samestyle\endcsname
\providecommand{\newblock}{\relax}
\providecommand{\bibinfo}[2]{#2}
\providecommand{\BIBentrySTDinterwordspacing}{\spaceskip=0pt\relax}
\providecommand{\BIBentryALTinterwordstretchfactor}{4}
\providecommand{\BIBentryALTinterwordspacing}{\spaceskip=\fontdimen2\font plus
\BIBentryALTinterwordstretchfactor\fontdimen3\font minus \fontdimen4\font\relax}
\providecommand{\BIBforeignlanguage}[2]{{%
\expandafter\ifx\csname l@#1\endcsname\relax
\typeout{** WARNING: IEEEtran.bst: No hyphenation pattern has been}%
\typeout{** loaded for the language `#1'. Using the pattern for}%
\typeout{** the default language instead.}%
\else
\language=\csname l@#1\endcsname
\fi
#2}}
\providecommand{\BIBdecl}{\relax}
\BIBdecl

\bibitem{Vaswani2017AttentionIA}
A.~Vaswani \emph{et~al.}, ``Attention is all you need,'' in \emph{Neural Information Processing Systems}, 2017.

\bibitem{7485869}
S.~Ren, K.~He, R.~Girshick, and J.~Sun, ``Faster {R}-{CNN}: {Towards} {Real}-{Time} {Object} {Detection} with {Region} {Proposal} {Networks},'' \emph{IEEE Trans. Pattern Anal. Mach. Intell.}, vol.~39, no.~6, pp. 1137--1149, 2017.

\bibitem{tian2025yolov12}
Y.~Tian, Q.~Ye, and D.~Doermann, ``Yolov12: Attention-centric real-time object detectors,'' \emph{arXiv preprint arXiv:2502.12524}, 2025.

\bibitem{Carion2020EndtoEndOD}
N.~Carion, F.~Massa, G.~Synnaeve, N.~Usunier, A.~Kirillov, and S.~Zagoruyko, ``End-to-end object detection with transformers,'' \emph{ArXiv}, vol. abs/2005.12872, 2020.

\bibitem{10242028}
C.~Chen, W.~Zeng, X.~Zhang, and Y.~Zhou, ``{CS}\(^n\){Net}: {A} {Remote} {Sensing} {Detection} {Network} {Breaking} the {Second}-{Order} {Limitation} of {Transformers} {With} {Recursive} {Convolutions},'' \emph{IEEE Trans. Geosci. Remote Sens.}, vol.~61, pp. 1--15, 2023.

\bibitem{Chen2024RSNetAL}
H.~Chen, C.~Chen, F.~Wang, Y.~Shi, and W.~Zeng, ``Rsnet: A light framework for the detection of multi-scale remote sensing targets,'' \emph{ArXiv}, vol. abs/2410.23073, 2024.

\bibitem{Yang2024Qwen25TR}
Q.~A. Yang \emph{et~al.}, ``Qwen2.5 technical report,'' \emph{ArXiv}, vol. abs/2412.15115, 2024.

\bibitem{Achiam2023GPT4TR}
``Gpt-4 technical report,'' 2023.

\bibitem{DeepSeekAI2025DeepSeekR1IR}
``Deepseek-r1: Incentivizing reasoning capability in llms via reinforcement learning,'' \emph{ArXiv}, vol. abs/2501.12948, 2025.

\bibitem{Bai2025Qwen25VLTR}
S.~Bai \emph{et~al.}, ``Qwen2.5-vl technical report,'' 2025.

\bibitem{10547418}
W.~Zhang, M.~Cai, T.~Zhang, Y.~Zhuang, and X.~Mao, ``Earthgpt: A universal multimodal large language model for multisensor image comprehension in remote sensing domain,'' \emph{IEEE Transactions on Geoscience and Remote Sensing}, vol.~62, pp. 1--20, 2024.

\bibitem{Touvron2023Llama2O}
H.~Touvron \emph{et~al.}, ``Llama 2: Open foundation and fine-tuned chat models,'' \emph{ArXiv}, vol. abs/2307.09288, 2023.

\bibitem{DBLP:journals/corr/abs-2010-11929}
A.~Dosovitskiy \emph{et~al.}, ``An image is worth 16x16 words: Transformers for image recognition at scale,'' \emph{CoRR}, vol. abs/2010.11929, 2020.

\bibitem{10555327}
Y.~Wang and P.~Ghamisi, ``Rsadapter: Adapting multimodal models for remote sensing visual question answering,'' \emph{IEEE Transactions on Geoscience and Remote Sensing}, vol.~62, pp. 1--13, 2024.

\bibitem{10738390}
W.~Zhang, M.~Cai, T.~Zhang, G.~Lei, Y.~Zhuang, and X.~Mao, ``Popeye: A unified visual-language model for multisource ship detection from remote sensing imagery,'' \emph{IEEE Journal of Selected Topics in Applied Earth Observations and Remote Sensing}, vol.~17, pp. 20\,050--20\,063, 2024.

\bibitem{rs13183690}
T.~Zhang \emph{et~al.}, ``{SAR} {Ship} {Detection} {Dataset} ({SSDD}): {Official} {Release} and {Comprehensive} {Data} {Analysis},'' \emph{Remote Sens.}, vol.~13, no.~18, 2021.

\bibitem{9127939}
S.~Wei, X.~Zeng, Q.~Qu, M.~Wang, H.~Su, and J.~Shi, ``{HRSID}: {A} {High}-{Resolution} {SAR} {Images} {Dataset} for {Ship} {Detection} and {Instance} {Segmentation},'' \emph{IEEE Access}, vol.~8, pp. 120\,234--120\,254, 2020.

\bibitem{CHENG2014119}
G.~Cheng, J.~Han, P.~Zhou, and L.~Guo, ``Multi-class geospatial object detection and geographic image classification based on collection of part detectors,'' \emph{ISPRS Journal of Photogrammetry and Remote Sensing}, vol.~98, pp. 119--132, 2014.

\bibitem{Jain2023NEFTuneNE}
N.~Jain \emph{et~al.}, ``Neftune: Noisy embeddings improve instruction finetuning,'' \emph{ArXiv}, vol. abs/2310.05914, 2023.

\bibitem{hu2021loralowrankadaptationlarge}
E.~J. Hu, Y.~Shen, P.~Wallis, Z.~Allen-Zhu, Y.~Li, S.~Wang, L.~Wang, and W.~Chen, ``Lora: Low-rank adaptation of large language models,'' 2021.

\end{thebibliography}

\newpage

\vfill

\end{document}